\documentclass[acmtog]{acmart}

\usepackage{booktabs} 
\usepackage{amsmath}
\usepackage{mathtools}
\usepackage{caption}
\usepackage{enumitem}
\usepackage{float}
\usepackage{graphicx}

\usepackage[capitalize,noabbrev]{cleveref}
\usepackage{xspace}

\crefname{section}{Sec.}{Secs.}
\Crefname{section}{Section}{Sections}
\crefname{table}{Tab.}{Tabs.}
\Crefname{table}{Table}{Tables}
\crefname{figure}{Fig.}{Figs.}
\Crefname{figure}{Figure}{Figures}
\crefname{equation}{Eq.}{Eqs.}
\Crefname{equation}{Equation}{Equations}
\newcommand{\method}{LAR-Gen\xspace}
\usepackage[ruled]{algorithm2e} 

\SetAlFnt{\small}
\SetAlCapFnt{\small}
\SetAlCapNameFnt{\small}
\SetAlCapHSkip{0pt}

\let\oldtwocolumn\twocolumn
\renewcommand\twocolumn[1][]{
    \oldtwocolumn[{#1}{
    \begin{center} \includegraphics[width=0.88\textwidth]{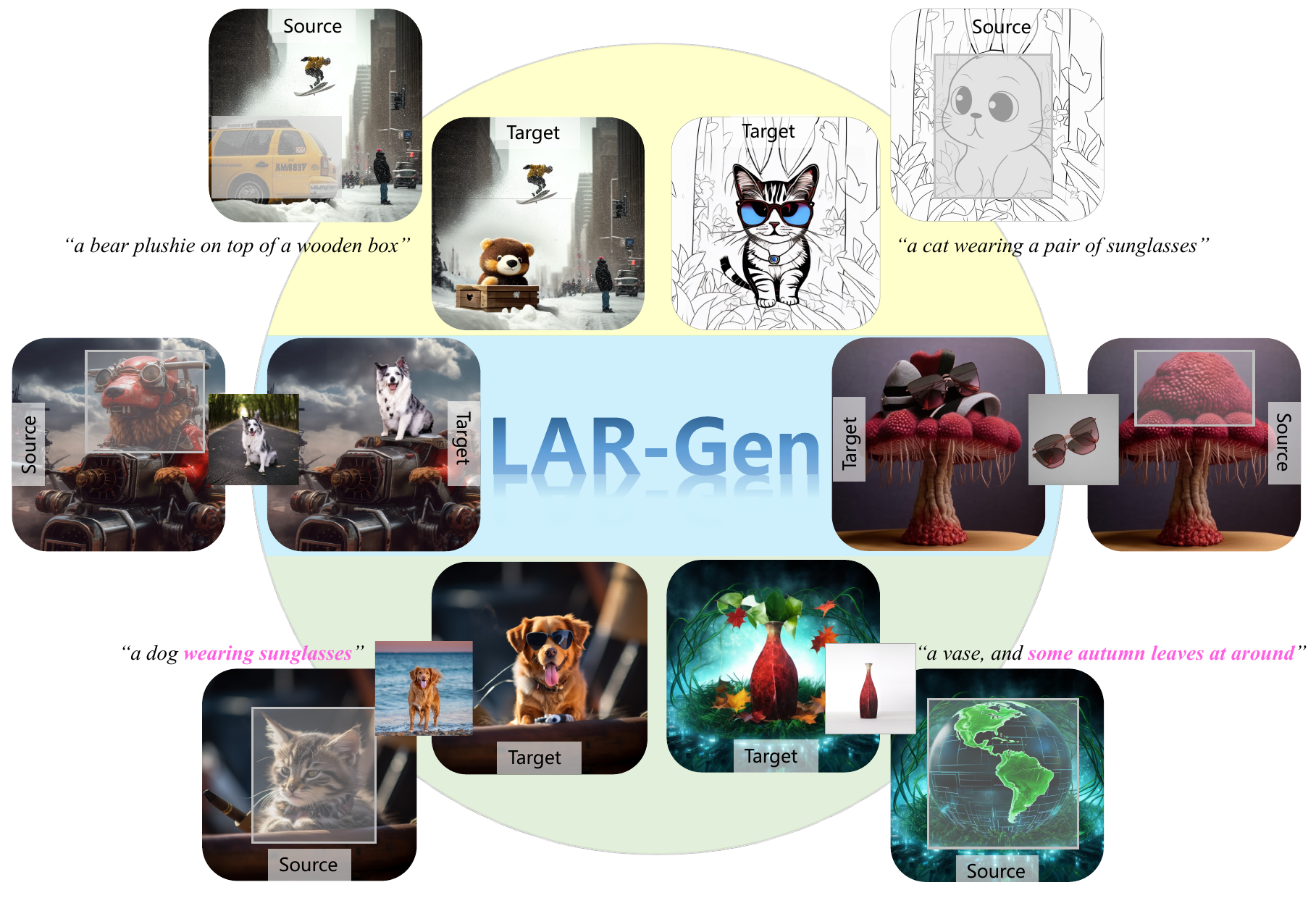}
        \captionof{figure}{Image inpainting results by our \method via multimodal prompt, i.e., mask prompt, image prompt, and text prompt. Given a source image, our \method could accurately inpaint the region of mask prompt. according to the guidance of \textit{text-only (i.e., text prompt, as shown in the top), image-only (i.e., image prompt, as shown in the middle), and their combination (as shown in the bottom)}. }
        \label{fig:teaser}
    \end{center}
    }]
}
\begin{document}
\title{Locate, Assign, Refine: Taming Customized Promptable Image Inpainting}

\author{Yulin Pan}
\affiliation{%
  \institution{Alibaba Group}
  \country{China}}
\email{yanwen.pyl@alibaba-inc.com}

\author{Chaojie Mao}
\affiliation{%
  \institution{Alibaba Group}
  \country{China}}
\email{chaojie.mcj@alibaba-inc.com}

\author{Zeyinzi Jiang}
\affiliation{%
  \institution{Alibaba Group}
  \country{China}}
\email{zeyinzi.jzyz@alibaba-inc.com}

\author{Zhen Han}
\affiliation{%
  \institution{Alibaba Group}
  \country{China}}
\email{hanzhen.hz@alibaba-inc.com}

\author{Jingfeng Zhang}
\affiliation{%
  \institution{Alibaba Group}
  \country{China}}
\email{zhangjingfeng.zjf@alibaba-inc.com}

\author{Xiangteng He}
\affiliation{%
  \institution{Wangxuan Institue of Computer Technology, Peking University}
  \country{China}}
\email{hexiangteng@gmail.com}

\renewcommand\shortauthors{Pan et al}

\begin{abstract}
Prior studies have made significant progress in image inpainting guided by either text description or subject image. However, the research on inpainting with flexible guidance or control, i.e., text-only, image-only, and their combination, is still in the early stage. Therefore, in this paper, we introduce the \textit{multimodal promptable image inpainting project: a new task model, and data for taming customized image inpainting}. We propose \textbf{\method}, a novel approach for image inpainting that enables seamless inpainting of specific region in images corresponding to the mask prompt, incorporating both the text prompt and image prompt. Our \method adopts a coarse-to-fine manner to ensure the \textit{context consistency} of source image, subject \textit{identity consistency}, local \textit{semantic consistency} to the text description, and \textit{smoothness consistency}. It consists of three mechanisms: (i) \textbf{Locate mechanism}: concatenating the noise with masked scene image to achieve precise regional editing, (ii) \textbf{Assign mechanism}: employing decoupled cross-attention mechanism to accommodate multi-modal guidance, and (iii) \textbf{Refine mechanism}: using a novel RefineNet to supplement subject details. Additionally, to address the issue of scarce training data, we introduce a novel data engine to automatically extract substantial pairs of data consisting of local text prompts and corresponding visual instances from a vast image data, leveraging publicly available pre-trained large models. Extensive experiments and various application scenarios demonstrate the superiority of \method in terms of both identity preservation and text semantic consistency.
%
\begin{figure*}[!t]
  \centering
  \includegraphics[width=\textwidth]{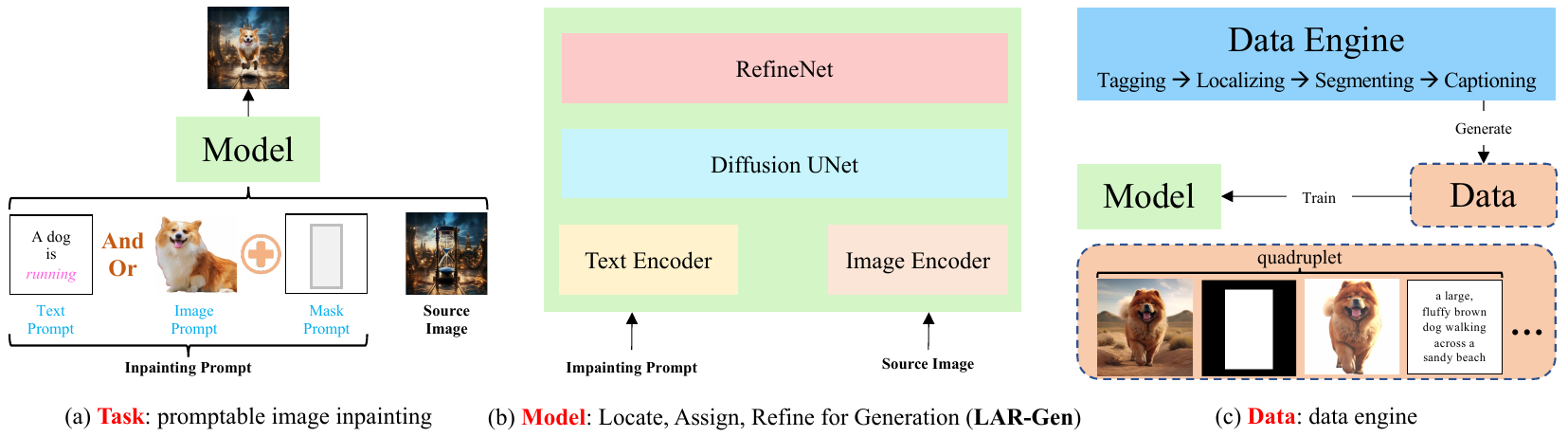}
  \caption{We build a unified framework for taming customized promptable image inpainting, consisting three interconnected components: a promptable image inpainting \textit{task}, a Locate, Assign, Refine for Generation \textit{model} (LAR-Gen), and a \textit{data} engine for constructing the data for training LAR-Gen.}
  \label{fig:TaskModelData}
\end{figure*}
\end{abstract}

%
%

%
%


\maketitle

\section{Introduction}
\label{sec:intro}

Thanks to the exhilarating advancements achieved by diffusion models~\cite{ddpm,glide,ldm,cfg}, image generation~\cite{midjourney,dalle3,wanx,pixart} is witnessing an exuberant proliferation and has found its application in various scenarios~\cite{controlnet,scedit,ip2p}.
Image inpainting, which aims to fill in missing regions of an image using various guiding inputs, has become as a significant application and recently demonstrated particularly attractive characteristics. 
Commonly, plain image inpainting technologies~\cite{lama,comodgan,madf} leverage the inherent semantics of the original image to appropriately fill in masked regions. 
Additionally, some auxiliary information could be incorporated to facilitate customized edits.
For example, text-guided image inpainting methods~\cite{sdinp,imageneditor} adjust images based on user-provided text prompt, as shown in the top of \cref{fig:teaser}. However, such methods often depend on global descriptions of the image, potentially limiting the fidelity of local semantic correspondence within the masked areas. 
Conversely, subject-guided image inpainting methods~\cite{anydoor,pbe,phd} reconstruct the masked region based on the referenced image prompt, as shown in the middle of \cref{fig:teaser}. Despite this, the prevalent use of a collage strategy~\cite{anydoor,phd} can result in issues resembling a simple copy-paste effect.
Besides, both text-only and image-only guided methods lack the precision required for fine control over the reconstruction of missing regions.

Therefore, in this paper, our goal is to build \textit{a unified framework for taming customized promptable image inpainting}, which means that we seek to develop a multimodal promptable model and train it on automatically constructed data for text and image flexible guided image inpainting, as shown in  \cref{fig:teaser}. It hingles on three components: \textbf{task, model, and data}, as shown in \cref{fig:TaskModelData}.

\textbf{TASK (Section \ref{sec:task}).} As shown in \cref{fig:TaskModelData} (a), we first introduce a new task that seamlessly integrates an arbitrary customized object (i.e., image prompt) into the desired location (i.e., mask prompt) within a source image, and allows for auxiliary text prompt to achieve fine-grained control.
This task is elaborately designed to accept a maximum input of quadruplet, including a source image, a mask prompt, an image prompt containing a subject, and a text prompt, adhering to the following distinct principles:
\begin{itemize}
\item 
\textbf{Context consistency.} The background area of the source image should remain unchanged, focusing solely on the editing in the specific area of  mask prompt.
\item 
\textbf{ID consistency.} The subject identity and details from image prompt should be preserved as much as possible.
\item 
\textbf{Semantic consistency.} The inpainted content should semantically correspond to the text prompt.
\item 
\textbf{Smoothness consistency.} The reconstructed image should achieve seamless integration with the context of the source image, and maintain high quality.
\end{itemize}

\textbf{MODEL (Section \ref{sec:method}).}
As shown in \cref{fig:TaskModelData} (b), we propose a unified promptable image inpainting model, termed as \textbf{\method}, which follows a ``\textbf{L}ocate, \textbf{A}ssign, \textbf{R}efine'' pipeline to achieve the above objectives and enable creative \textbf{Gen}eration. 
Specifically, the source image is first encoded into latent space and concatenated with the noise input, along with the mask prompt. This stage compels the model to seamlessly inpaint the masked region while keeping the background unaltered.
Then, a decoupled cross-attention mechanism is designed to effectively guide the diffusion process under the joint control of the text prompt and the image prompt, ensuring that the guidance process conforms to the semantics of the local textual description and the coarse-grained subject reference.
Finally, an auxiliary U-Net~\cite{unet} termed RefineNet is introduced to supplement subject details. 
It encodes the noisy subject image at each sampling step and injects the detail features into the main branch at multi-scale self-attention layers, facilitating the subject detail preservation.
With the help of the collected quadruple data (details in Section \ref{sec:data}) and the tailored pipeline, \method not only excels in text-subject-guided image inpainting task, but also supports text-only and image-only guided inpainting within the same framework, as illustrated in \cref{fig:teaser}. 
This indicates that our \method is a unified framework for image inpainting and can be employed in various scenarios.
Besides, extensive experiments demonstrate the superiority of \method in terms of both subject identity consistency and text semantic consistency.

\textbf{DATA (Section \ref{sec:data}).}  \cref{fig:TaskModelData} (c), 
to address the scarcity of public datasets that provide subject images (i.e., image prompt) paired with localized textual descriptions (i.e., text prompt) for multimodal promptable image inpainting, we introduce an innovative data engine to automatically assemble the necessary quadruple data.
The pipeline operates by automatically tagging~\cite{ram}, localizing~\cite{groundingdino}, instance segmenting~\cite{sam}, and visual captioning~\cite{llava}, thereby producing region-level quadruples comprising a source image, a mask prompt, an image prompt and a text prompt.
\section{Related Work}
\label{sec:related}

\subsection{Text-to-image Generation} 
In the past few years, diffusion-based image generation~\cite{ddpm,ddim,sde,glide,blipdiff,blenddiff} has emerged as a burgeoning research trend. 
This is due to the diffusion models' superior ability to generate high-quality images compared with GANs~\cite{gan,stylegan} and auto-regressive models~\cite{cogview,vqgan}.
Large-scale diffusion models have been proven to be excellent starting points for various downstream tasks, such as image inpainting~\cite{sdinp,dreaminp,imageneditor}, and image super-resolution~\cite{stablesr,diffbir,pasd}. 
Our \method builds upon Stable Diffusion~\cite{ldm,sd15} to fully leverage its power on generating a high-fidelity image.

\subsection{Subject-driven Image Generation}
Subject-driven image generation~\cite{controlcom,instantid} aims to generate images conditioned on a customized subject and a text prompt that describes the context. 
Existing works can be categorized in terms of the necessity of test-time tuning. One line of them requires one more image to conduct training optimization for specific subjects. 
Textual Inversion~\cite{textinv} is the first work to inverse a subject into textual representation space. It fixes the U-Net~\cite{unet} backbone and only tunes the newly added embedding. 
DreamBooth~\cite{dreambooth} tunes the entire U-Net together with the registered embedding and introduces an auxiliary loss to prevent generative ability degradation. 
Custom Diffusion~\cite{customdiff} performs parameter updating only at cross-attention layers and is able to combine multiple subjects into one image. 
While they excel at capturing subject detail, the time-consuming nature of the tuning process renders these methods less efficient than zero-shot approaches, which in turn limits their practical application.
Another line of them encodes the subject with a visual encoder, avoiding test-time tuning. 
InstantBooth~\cite{instantbooth} and ELITE~\cite{elite} combine both global and local visual features and inject the fine-grained local features with an extra cross-attention layer to preserve the high-level details of the subject. 
CustomNet~\cite{customnet} proposes a novel view generation method for the subject, which introduces the camera extrinsic as an extra condition to control the generated view of the subject.
IP-Adapter~\cite{ipadapter} introduces a decoupled cross-attention mechanism, which achieves multi-modal prompts control by weighted mixing of two attention layer outputs.
Despite the significant advancements, these methods are still at the early stage of maintaining the subject's identity. Moreover, their inability to perform localized editing constrains their applicability.

\subsection{Image Inpainting}
Given a masked scene image, image inpainting aims to inpaint the masked region guided by visual or textual prompts.
While text-guided image inpainting technology~\cite{sdinp,imageneditor} has reached a stage of maturity, subject-driven image inpainting remains in its nascent phase.
Recently many works have been proposed to address subject-driven image inpainting. 
Paint-by-Exa~\cite{pbe} replaces the text embedding with global image embedding in the cross-attention layer to inject the subject information into the diffusion process. 
ObjectStitch~\cite{objsitch} introduces a content adapter to mapping the image embedding to textual embedding space and applies the mask to constrain the denoising area, thereby achieving image conditional editing. 
AnyDoor~\cite{anydoor} and PhD~\cite{phd} apply a ControlNet~\cite{controlnet} to control the diffusion process with a subject-scene collage. 
%
Although these approaches have achieved significant advancements in subject- or text-conditioned image inpainting, none of them supports multi-modal prompt control. 
This paper addresses the text-subject-guided image inpainting task, which enables the joint control of visual and textual prompts.
\section{Promptable Image Inpainting Task}
\label{sec:task}
The new promptable image inpainting task is defined as the source image is edited based on the multimodal prompts, i.e., mask prompt, image prompt, and text prompt, as shown in \cref{fig:TaskModelData} (a). 
Specifically, given a source image $x_{s} \in \mathbb{R}^{H \times W \times 3}$, a mask prompt $m \in \mathbb{R}^{H \times W}$, an image prompt $x_{\text{obj}}$ and a text prompt $s$, \method aims to inpaint a local region of $x_{s}$ specified by $m$, under the control of $x_{\text{obj}}$ and $s$ separately or jointly.
The inpainted content inside region should remain contextual area of the source iamge unchanged (i.e., \textit{context consistency}), preserve the subject identity referring to image prompt (i.e., \textit{ID consistency}), align with the semantic of text prompt (i.e., \textit{semantic consistency}), and achieve high-quality and seamless integration with the source image (i.e., \textit{smoothness consistency}).
\section{LAR-Gen: Promptable Image Inpainting Model}
\label{sec:method}

\begin{figure}[!t]
  \centering
  \includegraphics[width=\linewidth]{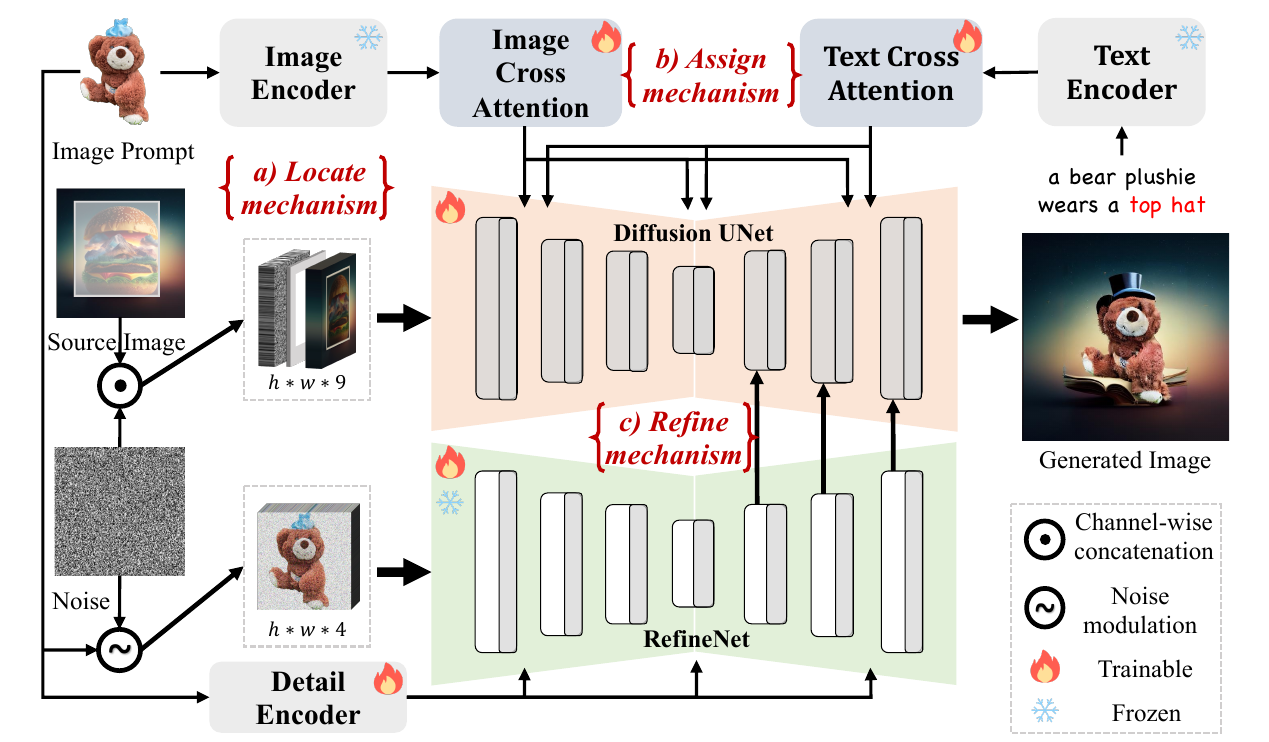}
  \caption{%
    Overall pipeline of \method.
  }
  \label{fig:framework}
\end{figure}

The overall pipeline of \method is depicted in \cref{fig:framework}, reconciling three mechanisms to achieve high-fidelity and seamless composition: \textbf{L}ocate, \textbf{A}ssign and \textbf{R}efine. a) \textit{Locate mechanism} concatenates the noise with source image and mask prompt to compel the model to seamlessly inpaint the masked region while keeping the background unaltered. b) \textit{Assign mechanism}, indeed a decoupled cross-attention mechanism, achieves text compatible subject guidance on denoising process. c) \textit{Refine mechanism}  gradually supplements the subject details, facilitating subject identity preservation.

\subsection{Locate Mechanism}
The text-to-image diffusion U-Net~\cite{unet} receives the latent code of a noisy image $z \in \mathbb{R}^{h \times w \times 4}$ and predicts the noise added to the image. 
To support local image editing, we concatenate the masked image with noise at channel dimension, following SD-Inpainting~\cite{sdinp}. It encodes the masked source image as:
\begin{equation}
    z_{s} = \text{Enc}(x_{s} \odot (1-m)),
\end{equation}
where $\text{Enc}$ represents the encoder of VAE of stable diffusion. $\odot$ represents the element-wise product operation.
Then we concatenate $z_{s}$ with $z$ and resized mask $m^{*} \in \mathbb{R}^{h\times w}$ to form the U-Net input:
\begin{equation}
    \Tilde{z} = [z; m^{*}; z_{s}],
\end{equation}
where $\Tilde{z} \in \mathbb{R}^{h \times w \times 9}$ and $[\cdot ; \cdot]$ denotes the concatenation operation.
Furthermore, we utilize a blend strategy during inference to ensure the background unaltered. It is formulated as:
\begin{equation}
    \hat{z}^{t} = z_{s}^{t} \odot (1 - m^{*}) + {z}^{t} \odot m^{*}
\end{equation}
where ${z}^{t}$ and $z_{s}^{t}$ denote the model output and the noisy latent code of source image at sampling step $t$, respectively. 

\subsection{Assign Mechanism}
While the pursuit of image inpainting conditioned independently on text descriptions or reference images has garnered considerable attention in recent years, mixing the two type of conditions presents a non-trivial challenge. 
Inspired by IP-adapter\cite{ipadapter}, we implement a decoupled cross-attention mechanism to enable compatible multi-modal prompt control.
Specifically, given the subject image prompt $x_\text{obj}$, we adopt a projection module $f$ attached to CLIP image encoder to encode the subject image prompt to a sequence of features $c_{i}$ with length $N$. It is formulated as:
\begin{equation}
    c_{i} = f(\text{CLIP}(x_{\text{obj}})),
\end{equation}
In our experiments, we adopt IP-Adapter-Plus\cite{ipadapter} as the projection module $f$ and freeze both CLIP encoder and the projection module during training. The feature sequence length $N$ is 16.
Subsequently, two extra parameters $\mathbf{W}_{k}^{'}$, $\mathbf{W}_{v}^{'}$ are introduced to insert image feature to cross-attention layer:
\begin{equation}
    \mathbf{K}^{'}=c_{i}\mathbf{W}_{k}^{'}, \mathbf{V}^{'}=c_{i}\mathbf{W}_{v}^{'},
\end{equation}
where $\mathbf{K^{'}, V^{'}}$ are the key and value matrices of the image-conditioned attention operation.
At each cross-attention layer, we first perform attention on image and text separately, which formulated as:
\begin{align}
    & \mathbf{Z} = \text{Softmax}(\frac{\mathbf{Q}\mathbf{K}^{\top}}{\sqrt{d}})\mathbf{V}, 
    & \mathbf{Z}^{'} = \text{Softmax}(\frac{\mathbf{Q}\mathbf{(K^{'})}^{\top}}{\sqrt{d}})\mathbf{V^{'}},
\end{align}
where $\mathbf{Z}$ and $\mathbf{Z}^{'}$ represent the text-conditioned and image-conditioned attention outputs, respectively. $\mathbf{Q, K, V}$ are the query, key, value matrices of the text-conditioned attention operation.
The final cross attention output is a weighted sum of $\mathbf{Z}$ and $\mathbf{Z}^{'}$:
\begin{equation}
    \mathbf{Z}^{\text{out}} = \mathbf{Z} + \beta\mathbf{Z}^{'},
\end{equation}
where $\beta$ is a hyper-parameter to modulate image control strength.

\subsection{Refine Mechanism}

To achieve text-subject-compatible guidance, the hyper-parameter $\beta$ in the decoupled cross-attention layer is generally set to a small value to ensure text consistency. 
However, this compromises image control and may result in the loss of subject details. 
To counteract this, we introduce an auxiliary U-Net, termed RefineNet, which enhances the object's details even when $\beta$ is small.
The RefineNet performs denoising process on the noisy subject image prompt, which is modulated with the same level of Gaussian noise. It utilizes a detail encoder, comprised of a CLIP image encoder and an MLP, to encode the subject's patch features:
\begin{equation}
    c_{i}^{'} = \text{MLP}(\text{CLIP}(x_{\text{obj}}))
\end{equation}
During training, the CLIP encoder is frozen and only the parameters of MLP are updated.
The feature $c_{i}^{'}$ is then used in the cross-attention layers of RefineNet to guide the diffusion process.
Hence the fine-grained details of subject can be captured by RefineNet at each sampling step $t$.
In each self-attention layer of the RefineNet's decoder, the input features that represent the subject details are stored.
These features are then concatenated with the corresponding layer's features of the main U-Net,  thereby encouraging the model to reconstruct the subject by leveraging both contextual information and detailed subject features. The self-attention operation in the main U-Net is formulated as:
\begin{equation}
    \mathbf{O}_{s} = \text{Softmax}(\frac{\mathbf{Q}_{s}\mathbf{K}_{s}^{\top}}{\sqrt{d}})\mathbf{V}_{s},
\end{equation}
where $\mathbf{Q}_{s}$, $\mathbf{K}_{s}$ and $\mathbf{V}_{s}$ represent the query, key and value matrices respectively. 
Given the context feature $c_{\text{ctx}}$ and the subject detail feature $c_{\text{obj}}$, the $\mathbf{Q}_{s}$, $\mathbf{K}_{s}$ and $\mathbf{V}_{s}$ can be calculated as:
\begin{equation}
    \mathbf{Q}_{s} = c_{\text{ctx}}\mathbf{W}_{qs}, \mathbf{K}_{s} = [c_{\text{ctx}}; c_{\text{obj}}]\mathbf{W}_{ks}, \mathbf{V}_{s} = [c_{\text{ctx}}; c_{\text{obj}}]\mathbf{W}_{vs}, 
\end{equation}
where $\mathbf{W}_{qs}$, $\mathbf{W}_{ks}$ and $\mathbf{W}_{vs}$ are the weight matrices of the self-attention layer in the main U-Net.

\begin{figure*}[t]
  \centering
  \includegraphics[width=\linewidth]{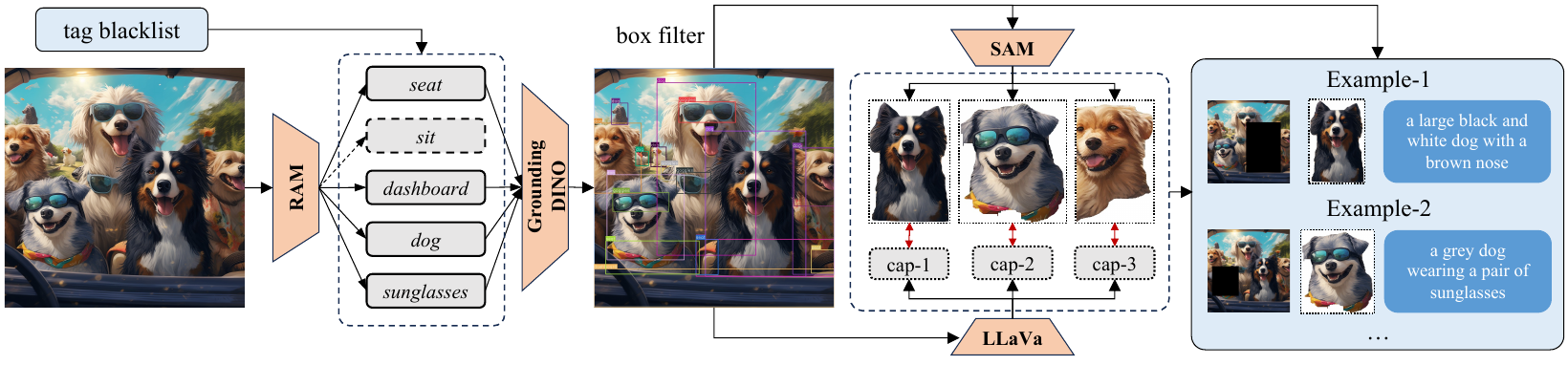}
  \caption{%
    Data engine, involving a sequence of automated processes including tagging, localizing, segmenting, and captioning. 
  }
  \label{fig:data_proc}
\end{figure*}

\subsection{Objective and Training}\label{subsec:training}

\method is trained with the mean square error loss that is a conventional choice among diffusion-family models. Given the concatenated latent code $\Tilde{z}$, the reference subject image prompt $x_{\text{obj}}$ and the text prompt $s$, the loss function is formulated as:
\begin{equation}
    \mathcal{L} = \mathbb{E}_{\Tilde{z},t, x_{\text{obj}},s,\epsilon \in \mathcal{N}(0, \mathbf{I})} \lVert \epsilon - \epsilon_{\theta}(\Tilde{z},t, x_{\text{obj}},s) \rVert_{2}
\end{equation}

A two-stage training strategy is adopted. Specifically, during the first stage, the RefineNet component is omitted, permitting focused training of the main U-Net, augmented by the locate and assign mechanism. 
The hyper-parameter $\beta$ is currently set to 1.
Upon transitioning to the secondary stage, $\beta$ is typically set to a constant value smaller than 1 to ensure the textual semantic alignment, while the RefineNet is integrated to supplement the object detail. 
At this stage, the main U-Net is frozen, with updates being confined solely to the parameters within the cross-attention layers of the RefineNet.

\subsection{Inference.} \label{subsec:inference}
We adopt classifier-free guidance\cite{cfg} during inference, a popular technique that has been proven effective in conditional image generation. 
Specifically, during training, both the image and text prompt are independently assigned a null value $\emptyset$ with a predetermined probability, thereby enabling the model to concurrently train on both conditional and unconditional denoising capabilities.
At inference, it shifts the score estimate towards the conditional direction and away from the unconditional direction:
\begin{equation}
    \Tilde{\epsilon}_{\theta}(\Tilde{z},t,x_{\text{obj}}, s) = \epsilon_{\theta}(\Tilde{z},t,\emptyset,\emptyset) + w(\epsilon_{\theta}(\Tilde{z},t,x_{\text{obj}}, s) - \epsilon_{\theta}(\Tilde{z},t,\emptyset,\emptyset)),
\end{equation}
where $w$ is the classifier-free guidance scale.

\section{Promptable Image Inpainting Data}
\label{sec:data}

To the best of our knowledge, there is no public dataset available that contains quadruplet data, including source image, mask prompt, image prompt, and text prompt. 
So we propose a data engine to automatically generate necessary training data via leveraging the power of pre-trained large models. It consists of four processing stages: tagging, localizing, segmenting, and captioning. 
Specifically, as depicted in \cref{fig:data_proc}, starting from a still image, we first apply RAM~\cite{ram} to extract the included tags within the image.
We exclude non-entity tags such as ``sky'', ``nature'' and ``skin''.
Grounding-DINO~\cite{groundingdino} and SAM~\cite{sam} are then utilized in turn to localize bounding box and segment mask region of each instance corresponding to each tag.
Objects too big or too small are also excluded. 

\begin{figure}[!t]
  \centering
  \includegraphics[width=\linewidth]{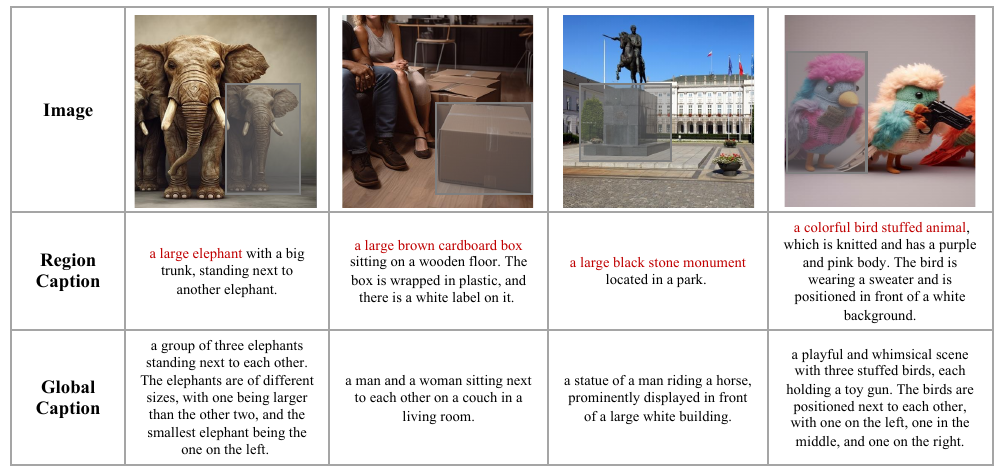}
  \caption{%
    Comparison between global caption and regional caption.
  }
  \label{fig:regioncap}
\end{figure}

The standard inpainting methods train the model with (image, prompt) pair data, where the prompt is a global description of image. 
However, we find that global caption may ignore the target subject or fail to describe it in detail. 
This leads to semantic misalignment on our training stage, and sometimes fail to generate subject. 
To address this, we crop the object region according to bounding box and send it to LLaVa~\cite{llava} to produce the regional caption. 
\cref{fig:regioncap} presents several examples to demonstrate the distinctions between global captions and region-specific captions.

\begin{figure}[!t]
  \centering
  \includegraphics[width=\linewidth]{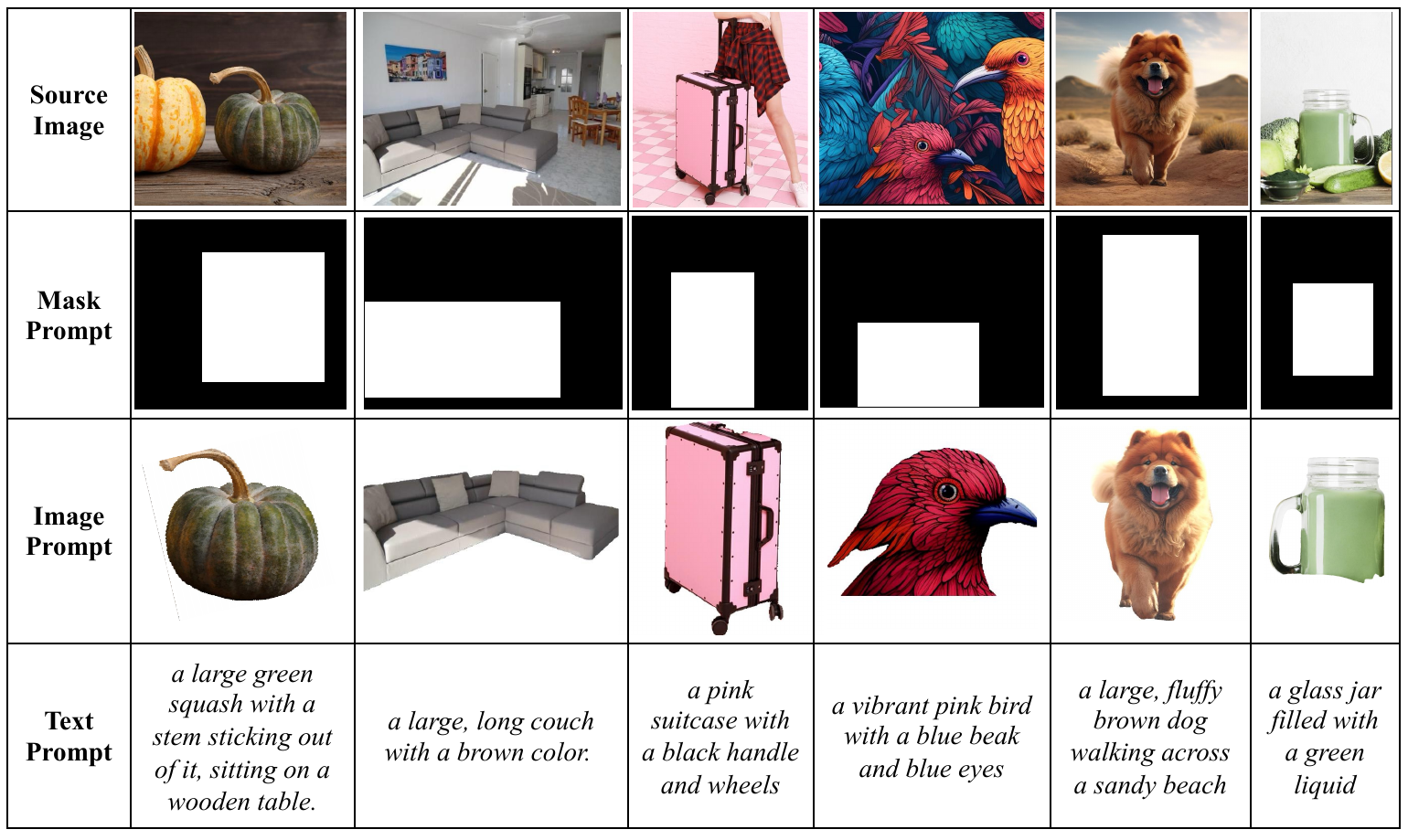}
  \caption{Visualization of some training samples.
  }
  \label{fig:traincase}
\end{figure}

By means of this process, we are able to extract quadruplets consisting of \textit{(source image, mask prompt, image prompt, text prompt)} from a vast repository of image data, as shown in \cref{fig:traincase}.

\section{Experiments}
\label{sec:exp}

\begin{figure*}[t]
  \centering
  \includegraphics[width=0.88\linewidth]{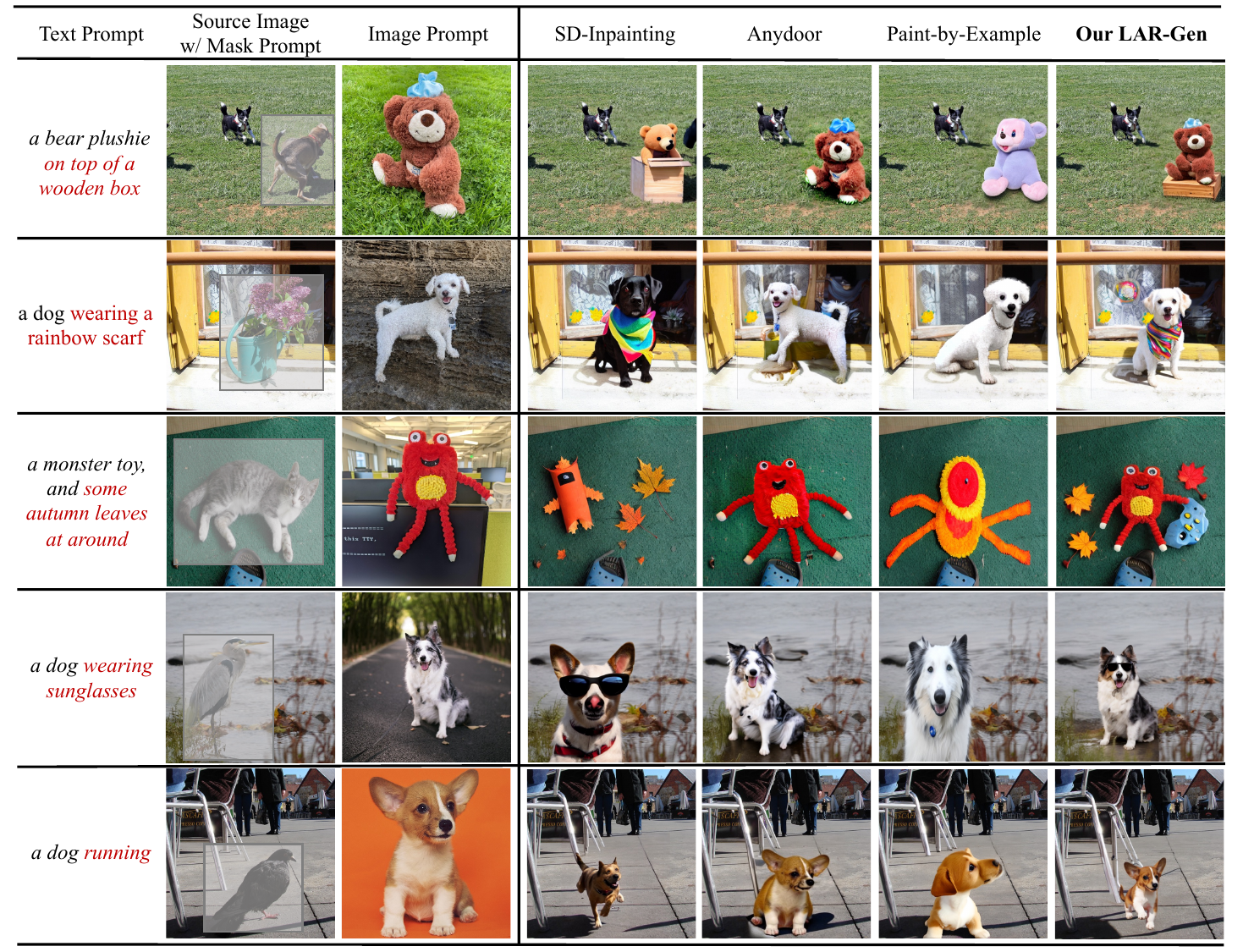}
  \caption{%
    Comparison with existing methods. Our \method achieves customized image inpainting corresponding to both text and image prompts, while others depend solely on text or image prompts.
  }
  \label{fig:sota}
\end{figure*}

\subsection{Dataset and Metrics}
 
\begin{figure}[H]
  \centering
  \includegraphics[width=\linewidth]{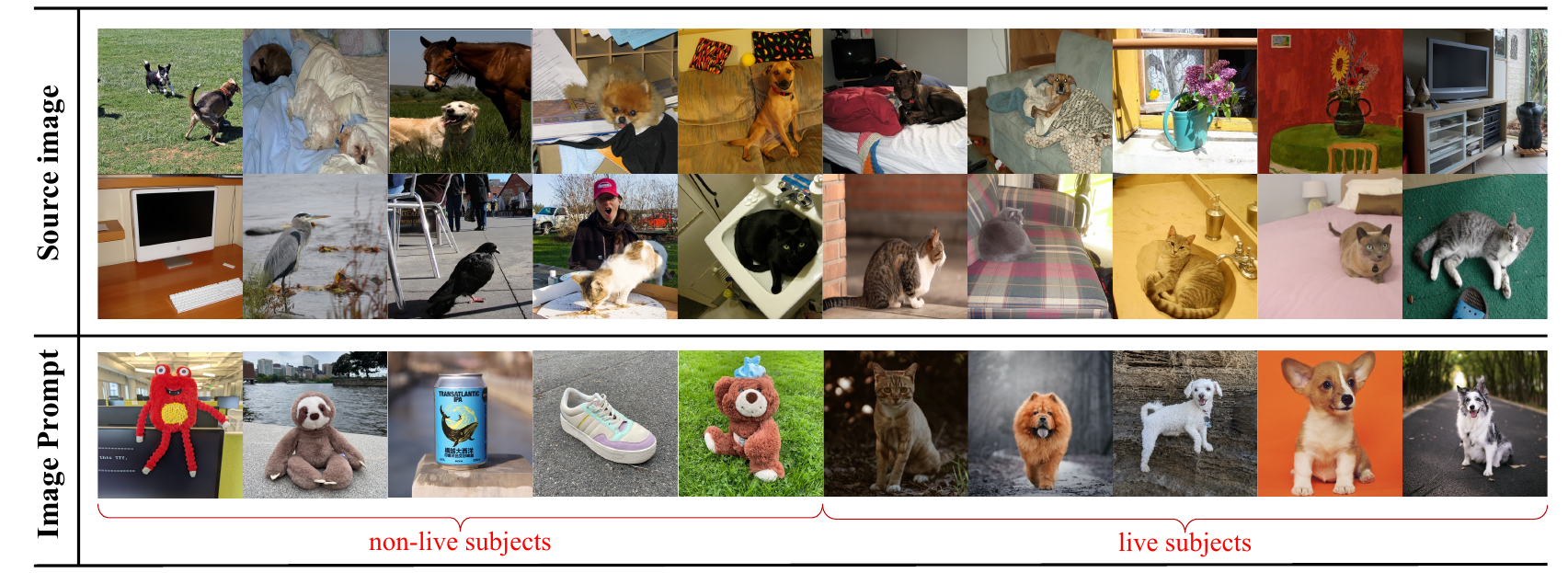}
  \caption{%
    Visualization of source images and image prompts for evaluation.
  }
  \label{fig:evaldata}
\end{figure}

\begin{figure}[H]
  \centering
  \includegraphics[width=\linewidth]{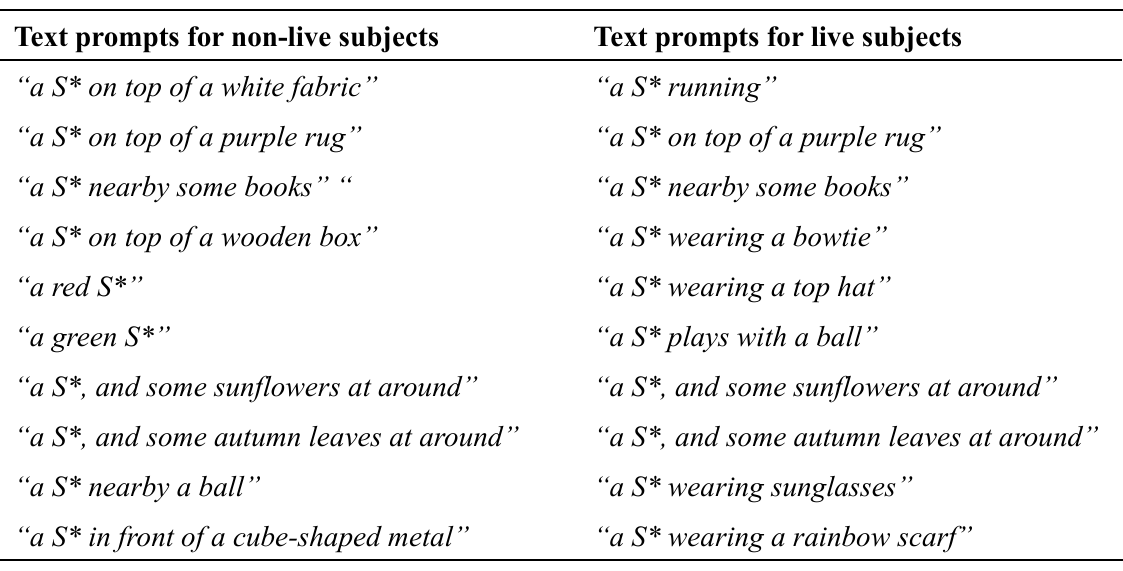}
  \caption{%
    Text prompt list for quantitative evaluation. 
  }
  \label{fig:textprompt}
\end{figure}

To evaluate the capability of our method on both subject identity consistency and text semantic consistency, we construct a benchmark that contains 2,000 \textit{(source image, mask prompt, image prompt, text prompt)} samples, using 20 scene images, 10 customized objects, and 10 pre-defined text prompts. 
The image prompts are provided by DreamBooth~\cite{dreambooth} dataset, and we manually pick 5 non-live subjects and 5 live subjects for fair comparison. 
For the source images and mask prompts, we randomly pick 20 images with corresponding bounding boxes from the COCO-val dataset~\cite{coco}. The source images and image prompts are shown in \cref{fig:evaldata}.
Then, for each \textit{(source image, mask prompt, image prompt)} pair, we bind it with 10 pre-defined text prompts (as shown in \cref{fig:textprompt}) to form the final test dataset.

Following Dreambooth, we evaluate the subject identity consistency by calculating the clip~\cite{clip} score between the inpainted region and the background-free subject image, denoted as CLIP-I~\cite{dreambooth}. 
Additionally, the text semantic consistency is also evaluated by calculating the clip score between the inpainted region and text prompt, denoted as CLIP-T~\cite{dreambooth}.


\begin{figure}[H]
  \centering
  \includegraphics[width=\linewidth]{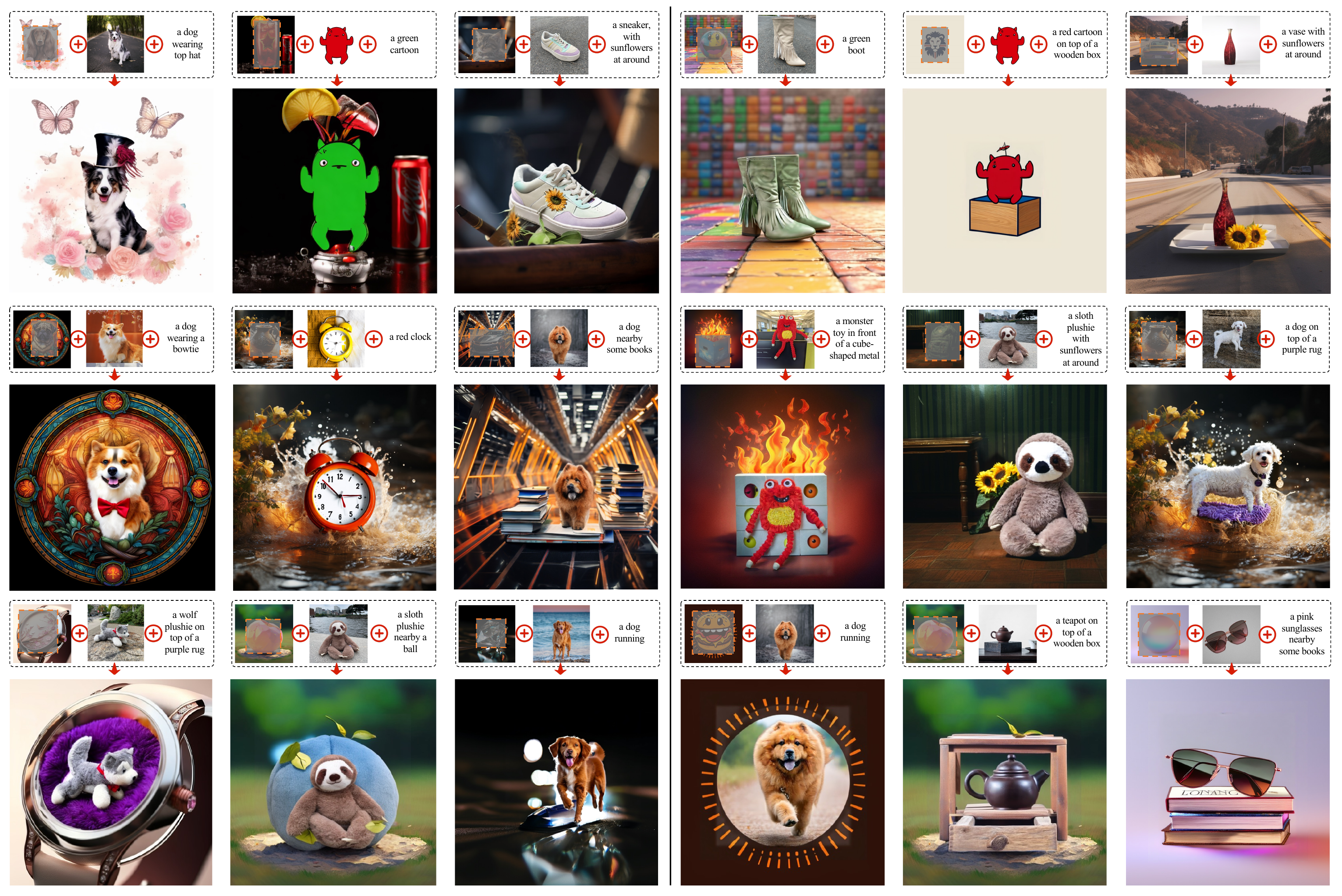}
  \caption{%
    Qualitative results of Stable Diffusion V1.5 (left) and Stable Diffusion XL (right).
  }
  \label{fig:sd15xl}
\end{figure}

\subsection{Implementation Details}
We apply Stable Diffusion V1.5~\cite{sd15} as the base architecture in our experiments, while other UNet-based architecture is also compatible with our \method, \textit{e.g.}, Stable Diffusion XL~\cite{sdxl}. Qualitative results for both architectures are shown in \cref{fig:sd15xl}.
We use the zoom-in strategy, which involves cropping a subimage from the source image around the masked region as the input for the diffusion model.
The model is trained at a batch size of 128 for 20k steps with a learning rate of 1e-5 for both two training stages.
The hyper-parameter $\beta$ is set to 1 during the first training stage, and adjusted to 0.3 for the second.
The classifier-free guidance scale $w$ is set to 7.5 during inference. 
All experiments are conducted on A100 gpu of 80G memory.

\subsection{Comparison with Existing Methods}

We first compare our \method with some existing inpainting methods, including SD-Inpainting~\cite{sdinp}, Paint-by-Exa~\cite{pbe}, and Anydoor~\cite{anydoor}. 
Note that SD-Inpainting supports only text prompt based condition, whereas Paint-by-Exa and Anydoor are limited to using a reference image prompt as the condition.
Unlike them, our \method is able to inpaint image by simultaneously leveraging the image and text prompts for joint guidance.
As depicted in \cref{fig:sota}, the qualitative outcomes demonstrate that our \method not only ensures that the inpainted region meets the semantic demands of the input prompt but also accurately aligns the generated content with that of the referenced image prompt.
\begin{table}[!t]
  \centering
  \caption{%
    Comparision with SOTA methods. Here $\beta$ represents the image prompt control strength, $s=\emptyset$ indicates that the text prompt is omitted.
  }
  \begin{small}
    \begin{tabular}{lccc}
    \toprule
    Methods & Prompts & CLIP-I & CLIP-T\\
    \midrule
    SD-Inpainting ~\cite{sdinp}   & text\&mask               & 68.68    & \textbf{30.37} \\
    Paint-by-Exa ~\cite{pbe}     &image\&mask          & 79.02    & 24.70 \\
    Anydoor ~\cite{anydoor}   &image\&mask                 & 82.10    & 25.52 \\
    \method ($\beta=0.3$)    &text\&image\&mask       & 79.15    & \textbf{29.94} \\
    \method ($\beta=1.0, s=\emptyset$) &text\&image\&mask & \textbf{83.39}    & 26.07 \\
    \bottomrule
    \end{tabular}
    \end{small}
  \label{tab:sota_comp}
\end{table}
We present the quantitative results in \cref{tab:sota_comp}.
We observe that when $\beta$ is set to 0.3, our method achieves a CLIP-T score comparable to that of SD-Inpainting, greatly surpassing Paint-by-Exa and Anydoor, thereby demonstrating its superior capability for maintaining semantic consistency with text prompt.
Furthermore, our \method surpasses SD-Inpainting and Paint-by-Exa by 10.47\% and 0.13\% in terms of CLIP-I score under $\beta=0.3$.
However, it trails Anydoor by a margin of 2.94\% in terms of CLIP-I scores, due to the fact that the generated object is supposed to be deformed to align the text prompt.
We also evaluate the performance of \method under the setting of $\beta=1.0$ and no text condition. \method achieves the highest CLIP-I score among all approaches, surpassing Anydoor by 1.29\%. 
This result demonstrates its superior capability for preserving the subject identity consistency with image prompt.

\begin{figure}[t]
  \centering
  \includegraphics[width=\linewidth]{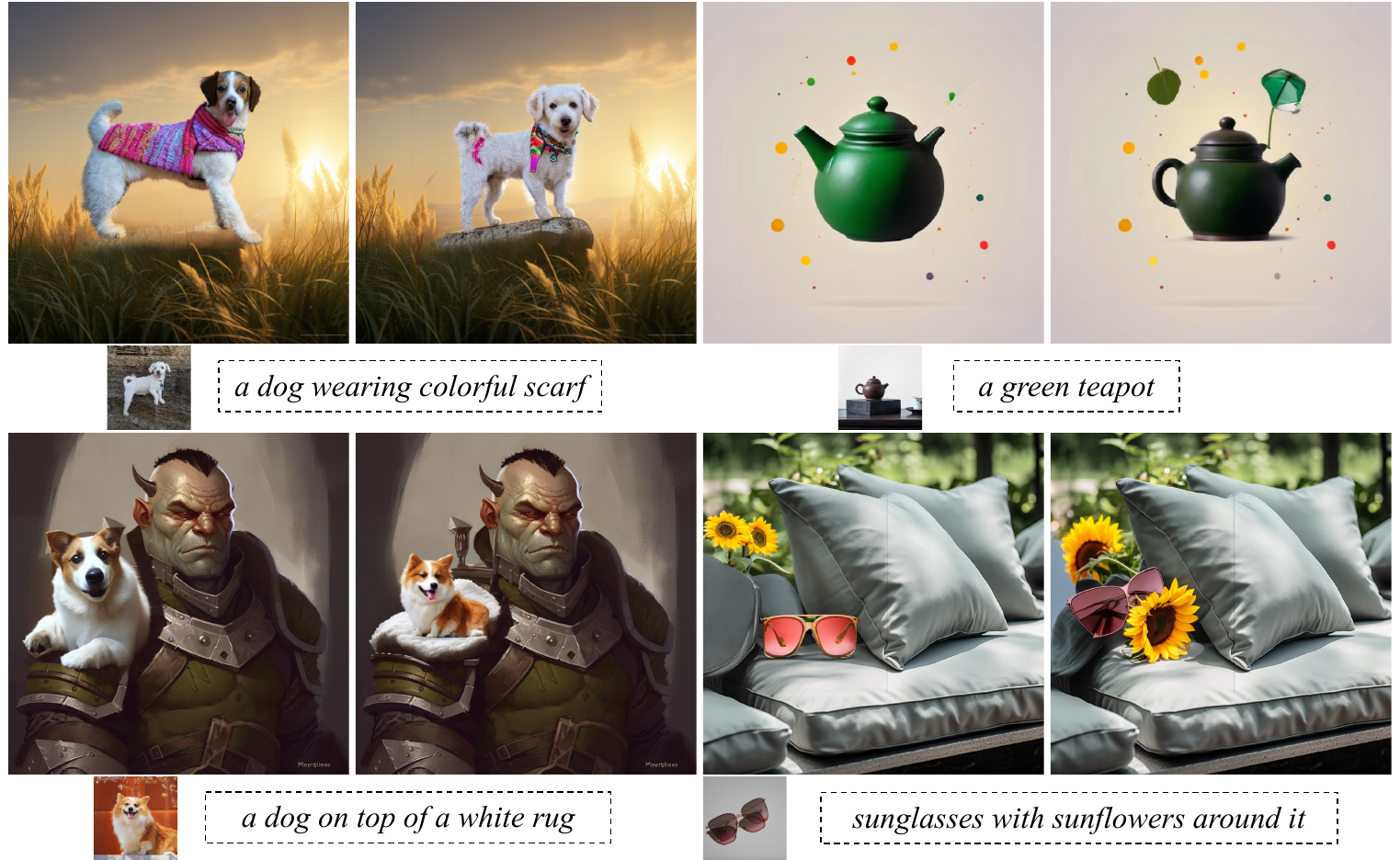}
  \caption{%
    Qualitative results of comparison with IP-Adapter+SD-Inpainting (shortened to SDIIPA). For each pair, the image on the left is generated by SDIIPA, while the right one is by our \method. 
  }
  \label{fig:IPSD}
\end{figure}

To further verify the superiority of our \method, we present a qualitative comparison with SD-Inpainting~\cite{sdinp}+IP-Adapter~\cite{ipadapter} (shortened to SDIIPA). The hyper-parameter $\lambda$ is set to 0.5 for SDIIPA (larger $\lambda$ causes the neglect of text) and 0.3 for our method. As shown in \cref{fig:IPSD}, for each pair, the image on the left is generated by SDIIPA, while the right one is by our \method. It can be observed that our \method consistently outperforms SDIIPA in terms of text semantic alignment and subject identity preservation.

\subsection{Ablation Studies}
\begin{table}[t]
  \centering
  \caption{%
    Ablation studies on core components of \method.
  }
  \setlength{\tabcolsep}{10pt}{
    \begin{tabular}{lcc}
    \toprule
    Module & CLIP-I & CLIP-T \\
    \midrule
    SD-Inpainting (Baseline)                  & 68.68    & 30.37 \\
    + Locate mechanism              & 68.60    & 31.01 \\
    ++ Assign mechanism             & 75.00    & 30.53 \\
    +++ Refine mechanism            & 79.15    & 29.94 \\
    \bottomrule
    \end{tabular}
  }
  \label{tab:abl_core}
\end{table}
\subsubsection{Effectiveness of Each Component}
We conduct experiments to verify the effectiveness of core components in our \method, i.e., Locate mechanism, Assign mechanism and Refine mechanism.
Starting with the vanilla SD-Inpainting model as baseline, we incrementally incorporate each mechanism and assess its impact on subject identity and text semantic consistencies by calculating the CLIP-I and CLIP-T scores. 
In this context, Locate mechanism specifically refers to the use of regional captions that are associated with particular image regions during training, as opposed to relying on global captions.
This is because the SD-Inpainting\cite{sdinp} model already includes the operation of concatenating masked source images.
As shown in \cref{tab:abl_core}, it yields a 0.65\% improvement in the CLIP-T score over the SD-Inpainting model, suggesting that region-specific captions enhance text semantic consistency due to their more accurate descriptions compared to global captions.
Upon this, the addition of Assign mechanism results in a significant improvement in the CLIP-I score (\textit{i.e.}, 68.60\% $\rightarrow$ 75.00\%) with only a marginal decrease in the CLIP-T score.
This suggests that Assign mechanism can effectively incorporate the condition from the referenced image prompt without substantially compromising text semantic consistency.
Lastly, the inclusion of the Refine mechanism leads to a 4.15\% increase in CLIP-I score, underscoring its exceptional efficacy in enhancing the high-frequency details of the subject.

\begin{figure}[t]
  \centering
  \includegraphics[width=\linewidth]{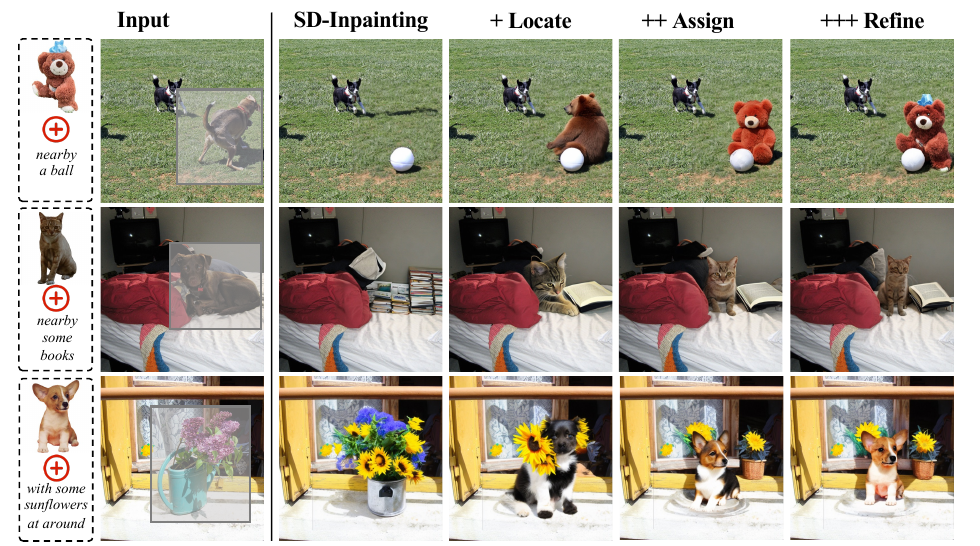}
  \caption{%
    Qualitative ablation studies on each proposed components. Starting with the standard SD-Inpainting model, we incrementally incorporate each component to evaluate its impact on performance.
  }
  \label{fig:abl_module}
\end{figure}

\subsubsection{Image Control Strength $\beta$}
\begin{figure}[ht]
  \centering
  \includegraphics[width=\linewidth]{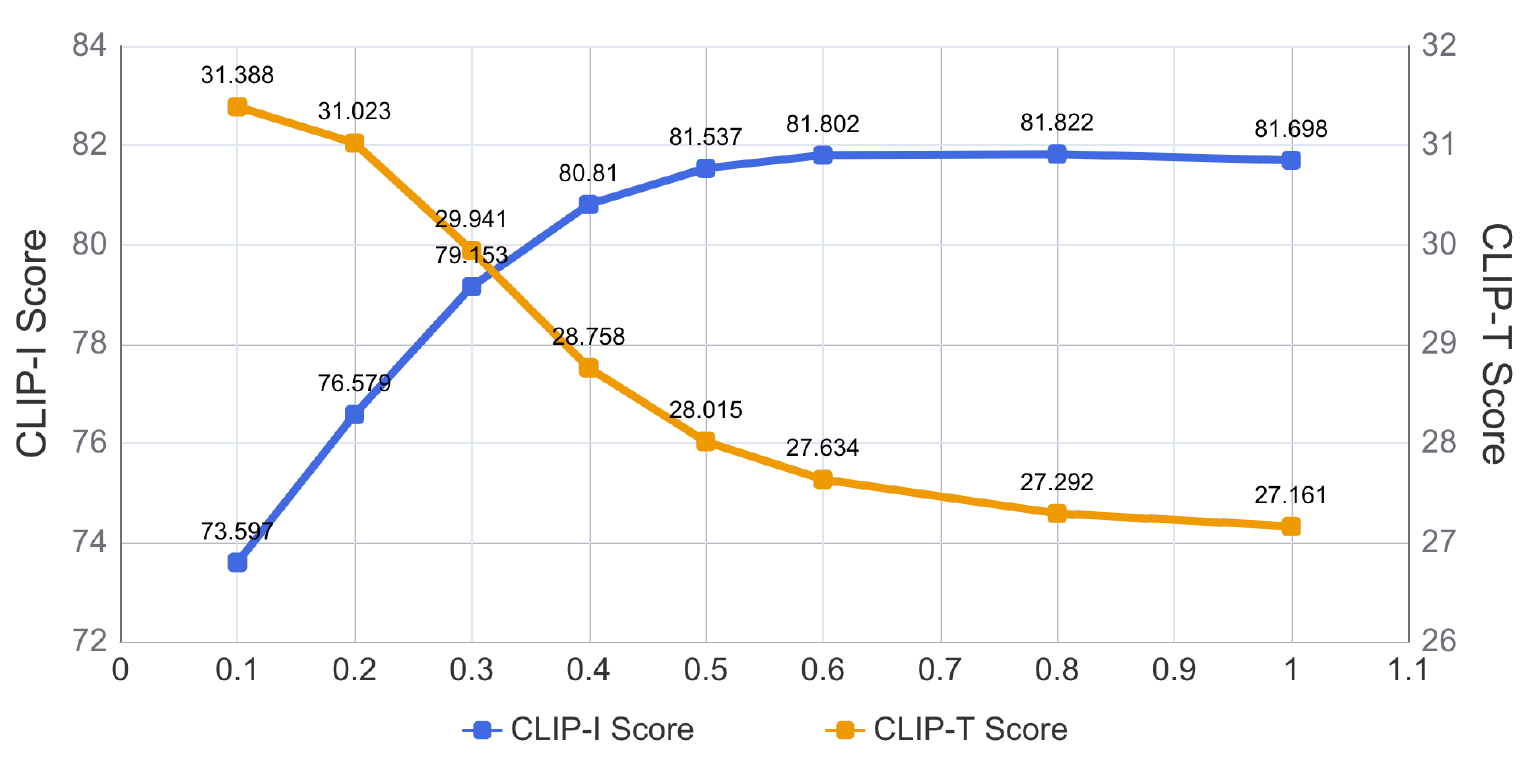}
  \caption{%
    Quantitative analysis of varying the value of hyper-parameter $\beta$.
  }
  \label{fig:beta_line}
\end{figure}

\begin{figure*}[t]
  \centering
  \includegraphics[width=0.88\linewidth]{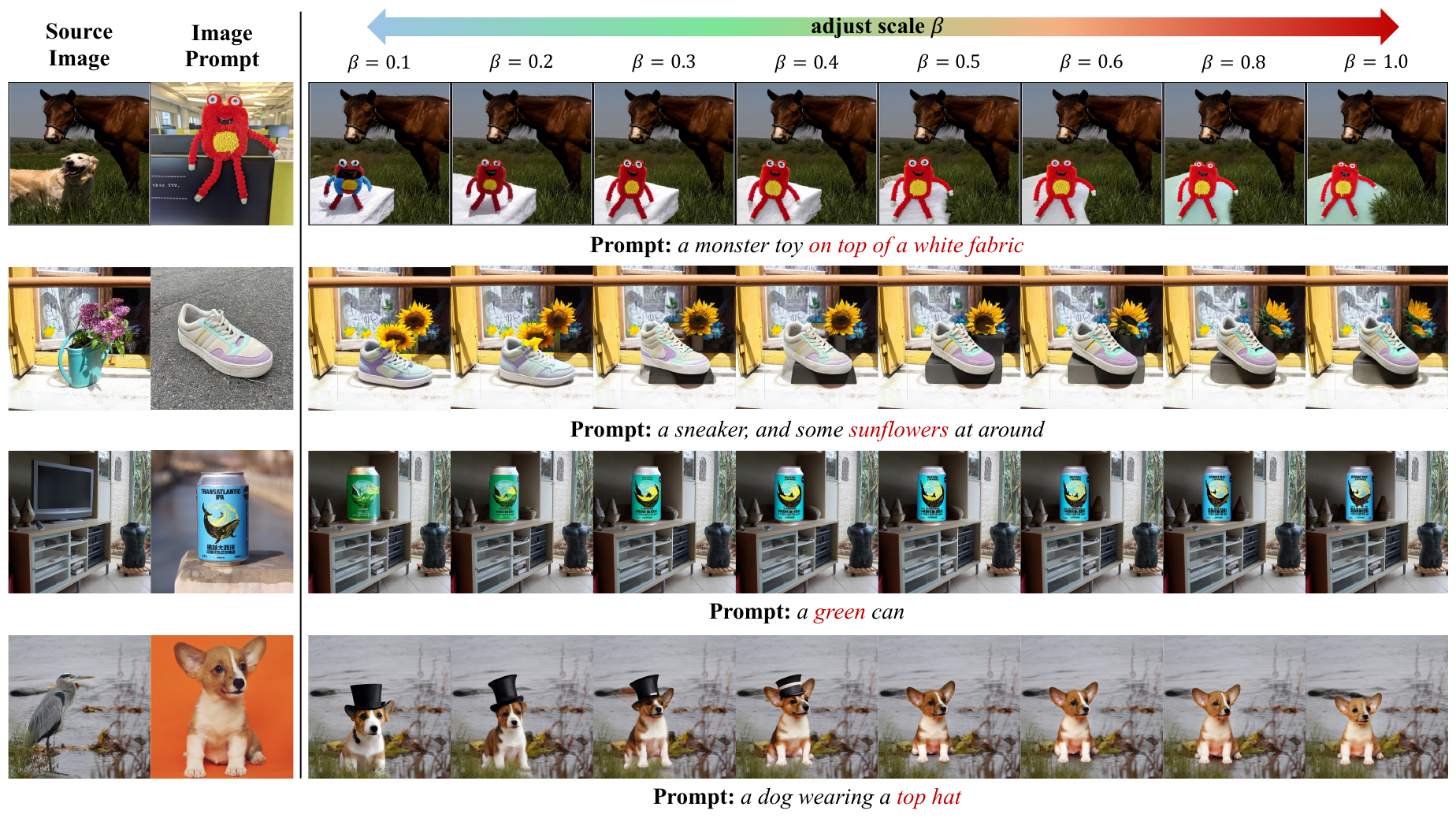}
  \caption{%
    Ablation studies on varying hyper-parameter $\beta$.
  }
  \label{fig:beta}
\end{figure*}

We conduct ablation experiments to clarify the contribution of hyper-parameter $\beta$. 
To this end, the first stage model is trained as described previously. 
Subsequently, in the second stage of training, we adjust $\beta$ from 0.1 to 1.0, which yields a series of second-stage models with varying control strengths.
We present quantitative results in \cref{fig:beta_line}, and visualization results in \cref{fig:beta}, respectively. 
From \cref{fig:beta_line}, we observe that the CLIP-I score progressively increases with the augmentation of $\beta$, while the CLIP-T score correspondingly diminishes. 
This observation corroborates the conclusion that $\beta$ plays a pivotal role in balancing the control between textual semantics and subject identity.
With the increase of scale $\beta$, we observe a greater retention of object details, albeit at the cost of reduced fidelity to the text prompts, as shown in \cref{fig:beta}.

\subsection{Diverse Applications}

\begin{figure*}[t]
  \centering
  \includegraphics[width=0.88\linewidth]{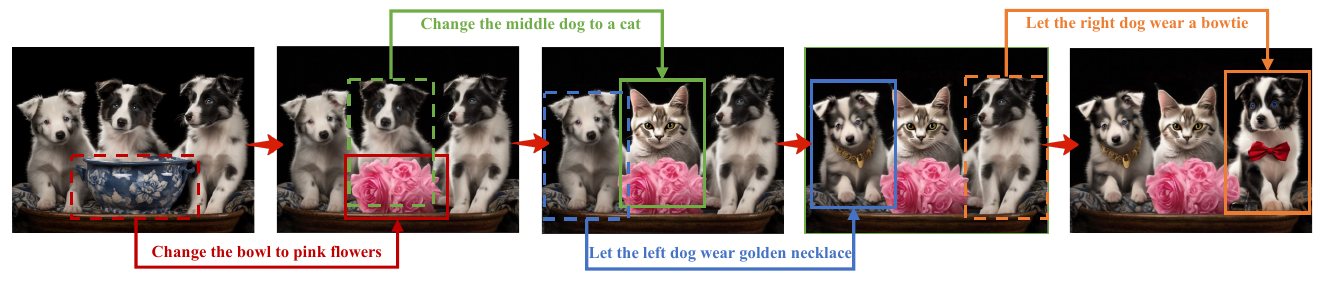}
  \caption{%
    Multi-turn and multi-type image editing results of our method. 
  }
  \label{fig:chain_edit}
\end{figure*}

\begin{figure}[H]
  \centering
  \includegraphics[width=\linewidth]{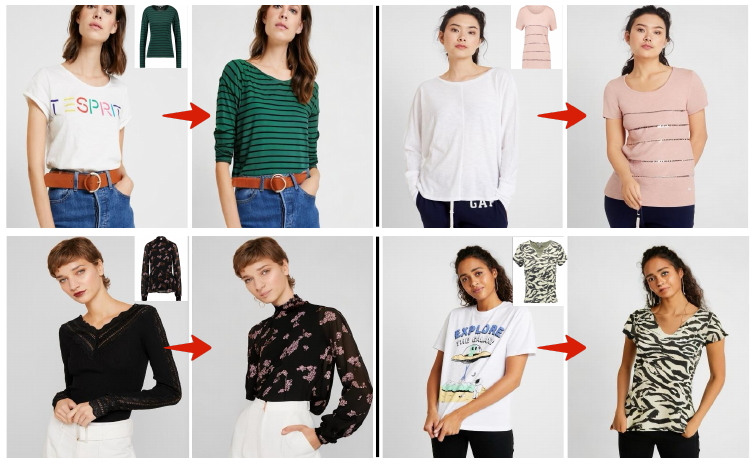}
  \caption{%
    Qualitative results on virtual try-on. The cases shown are selected from the test set of VITON-HD~\cite{vitonhd}.
  }
  \label{fig:cloth}
\end{figure}

Although \method is tailored for multimodal promptable image inpainting, it could serve as a unified image inpainting framework that supports text-only and image-only guided inpainting as well.
Benefiting from the unified framework, our \method could perform \textbf{multi-turn, multi-type inpainting}, as depicted in \cref{fig:chain_edit}, and show immense potential in a variety of applications, such as virtual try-on, object reshaping, instruct-based image editing, and so on.
We present several examples in \cref{fig:cloth} to illustrate its capabilities on \textbf{virtual try-on}, which randomly selected from the VITON-HD~\cite{vitonhd} test set. 
Note that our model has not been carefully optimized with try-on data. It requires only a single clothing image as input and does not need human pose or any other auxiliary information that generally required on traditional try-on methods.
Our \method achieves seamless cloth composition while preserving fine details of the clothing, such as color, texture, and pattern. 

\begin{figure}[H]
  \centering
  \includegraphics[width=\linewidth]{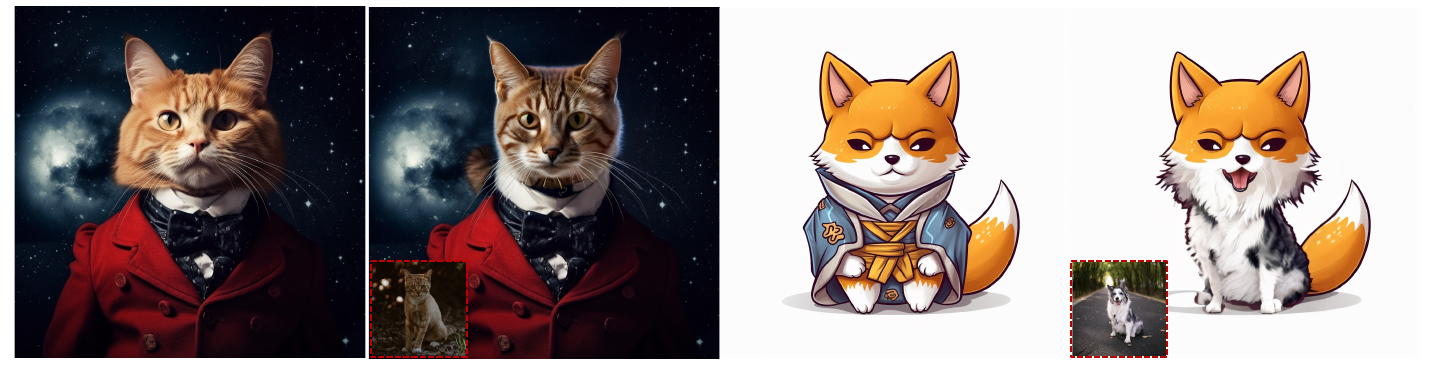}
  \caption{%
    The ability on inpainting object part.
  }
  \label{fig:partinpainting}
\end{figure}

Besides, Our \method is capable of \textbf{inpainting parts of objects} based on both the subject image and the context. We present some qualitative results to demonstrate this capability, as shown in \cref{fig:partinpainting}.

\section{Conclusion}

This paper proposes \method, a diffusion-based multimodal promptable image inpainting framework that could generate high-fidelity image with the joint guidance of text and subject image.
It follows a coarse-to-fine pipeline, which first generates the target subject by introducing the image embedding into the decoupled cross-attention mechanism, then refines the subject details with the proposed RefineNet, avoiding the copy-paste issue caused by collage-based strategy.
A hyper-parameter $\beta$ is introduced to balance the strength of image and text condition, thereby enabling fine-grained tuning to seek the optimal setting of preserving both subject identity and text semantic consistency.

Nevertheless, our \method still suffers from two limitations. 
Firstly, the reliance on a single subject image prompt as the conditional input may lead to inaccuracies in subject deformation. 
This is primarily due to the inherent ambiguity associated with generating deformations that have not been encountered previously, rendering the task inherently ill-posed. 
Secondly, the denoising process is influenced by multiple conditional factors, which may lead to the neglect of certain conditions, particularly when they conflict with others.
%
%

\bibliographystyle{ACM-Reference-Format}
\bibliography{sample-bibliography}

\end{document}


\title{Locate, Assign, Refine: Taming Customized Image Inpainting with Text-Subject Guidance \\
       \textit{- Supplementary Material -}
}

\maketitle

\appendix

\section{Training and Evaluation Data}
\begin{figure}[h]
  \centering
  \vspace{-3mm}
  \includegraphics[width=0.9\linewidth]{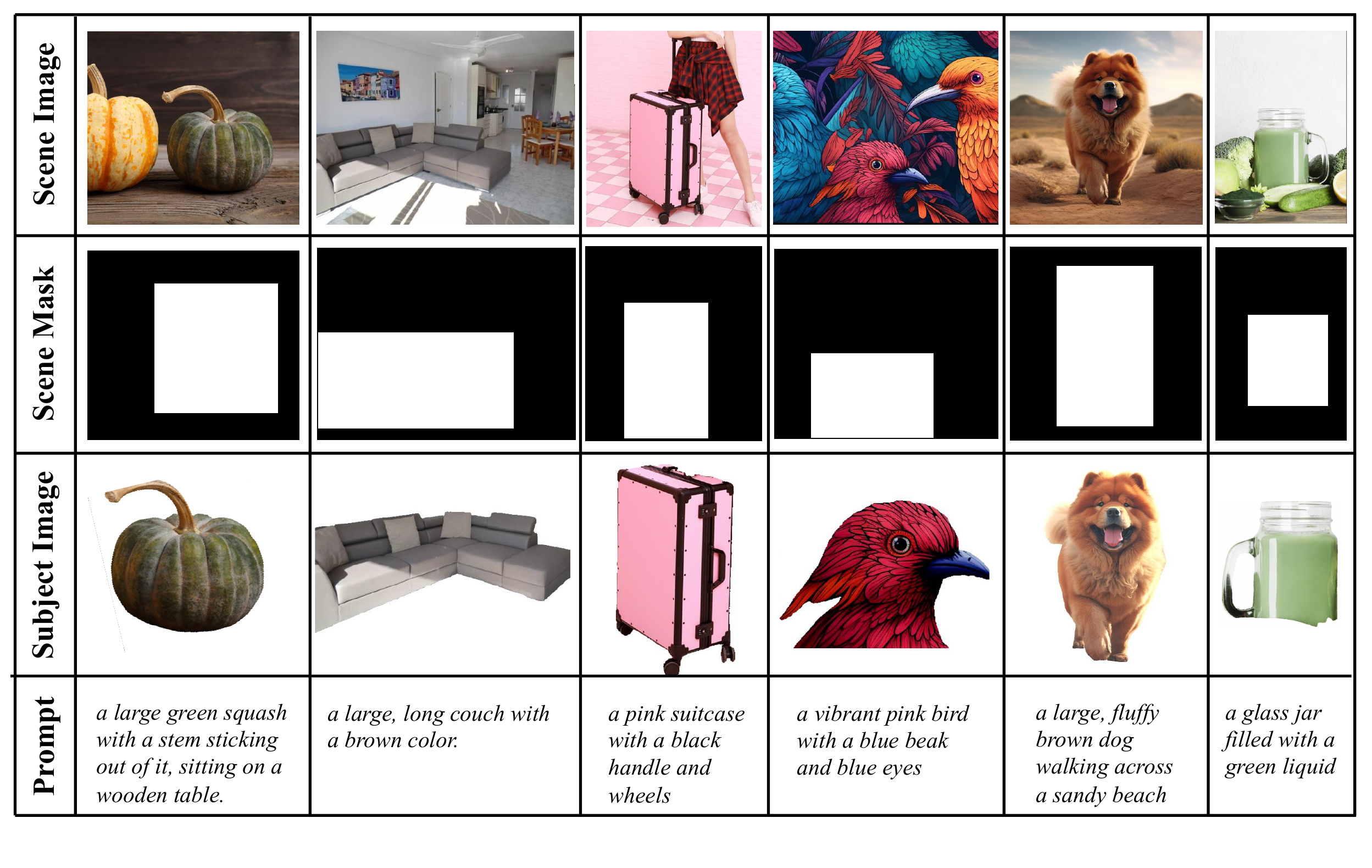}
  \vspace{-5mm}
  \caption{%
    \textbf{Visualization of some training samples.} Note that these are produced by the proposed data construction strategy.
  }
  \label{fig:train_case}
  \vspace{-5mm}
\end{figure}

In this section, we showcase some training samples that are generated using our proposed data construction strategy.
This strategy can create quadruplet data consisting of a scene image, a scene mask, a subject image, and a text prompt from a vast repository of image data. 
~\cref{fig:train_case} displays examples of these generated quadruplets.
The caption generated for the region accurately details the subject depicted in the subject image, demonstrating its effectiveness in terms of regional semantic alignment.
%

\begin{figure}[h]
  \centering
  \includegraphics[width=\linewidth]{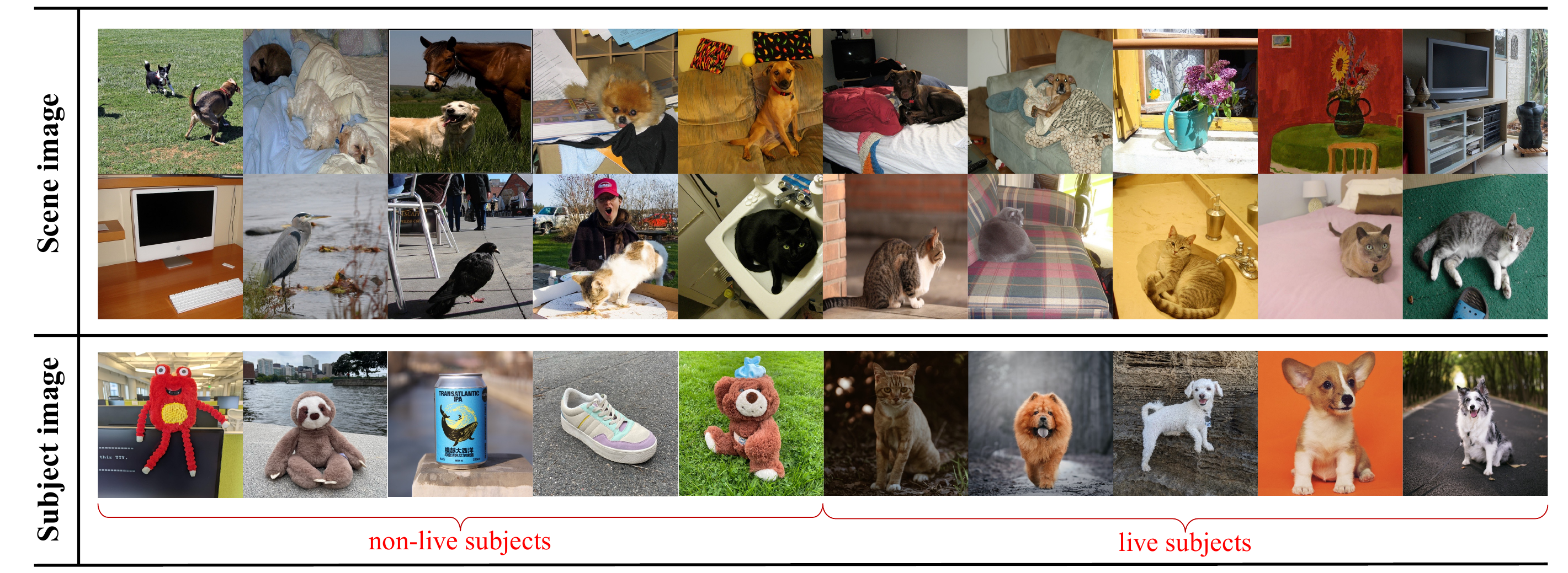}
  \caption{%
    \textbf{Visualization of scene images and subject images for evaluation.} The 20 scene images are randomly selected from COCO-val dataset and the 10 subjects are manually selected from DreamBooth dataset, containing 5 non-live subjects and 5 live subjects.
  }
  \label{fig:eval_image}
  \vspace{-0mm}
\end{figure}

Additionally, we present the scene and subject images used during the evaluation stage, as shown in \cref{fig:eval_image}. 
The 20 scene images are randomly selected from COCO-val\cite{coco} dataset and the 10 subjects are manually selected from DreamBooth\cite{dreambooth} dataset, containing 5 non-live subjects and 5 live subjects.
The textual descriptions are also presented in \cref{tab:prompt}. Following DreamBooth, we design different prompts for non-live and live subjects. Each subject is associated with 10 prompts.

\begin{table}[ht]
  \centering
  \caption{%
    \textbf{Text prompt list for quantitative evaluation.} At inference time, the placeholder S* is replaced with the category label of the specific subject.
  }
  \setlength{\tabcolsep}{2pt}{
    \begin{tabular}{ll}
    \toprule
    Text prompts for non-live subjects & Text prompts for live subjects \\
    \midrule
    ``a S* on top of a white fabric'' & ``a S* running'' \\
    ``a S* on top of a purple rug'' & ``a S* on top of a purple rug'' \\
    ``a S* nearby some books'' & ``a S* nearby some books'' \\
    ``a S* on top of a wooden box'' & ``a S* wearing a bowtie'' \\
    ``a red S*'' & ``a S* wearing a top hat'' \\
    ``a green S*'' & ``a S* plays with a ball'' \\
    ``a S*, and some sunflowers at around'' & ``a S*, and some sunflowers at around'' \\
    ``a S*, and some autumn leaves at around'' & ``a S*, and some autumn leaves at around'' \\
    ``a S* nearby a ball'' & ``a S* wearing sunglasses'' \\
    ``a S* in front of a cube-shaped metal'' & ``a S* wearing a rainbow scarf'' \\
    \bottomrule
    \end{tabular}
  }
  \vspace{-0mm}
  \label{tab:prompt}
\end{table}

\section{Qualitative Results of Different Backbones}
\begin{figure}[t]
  \centering
  \includegraphics[width=\linewidth]{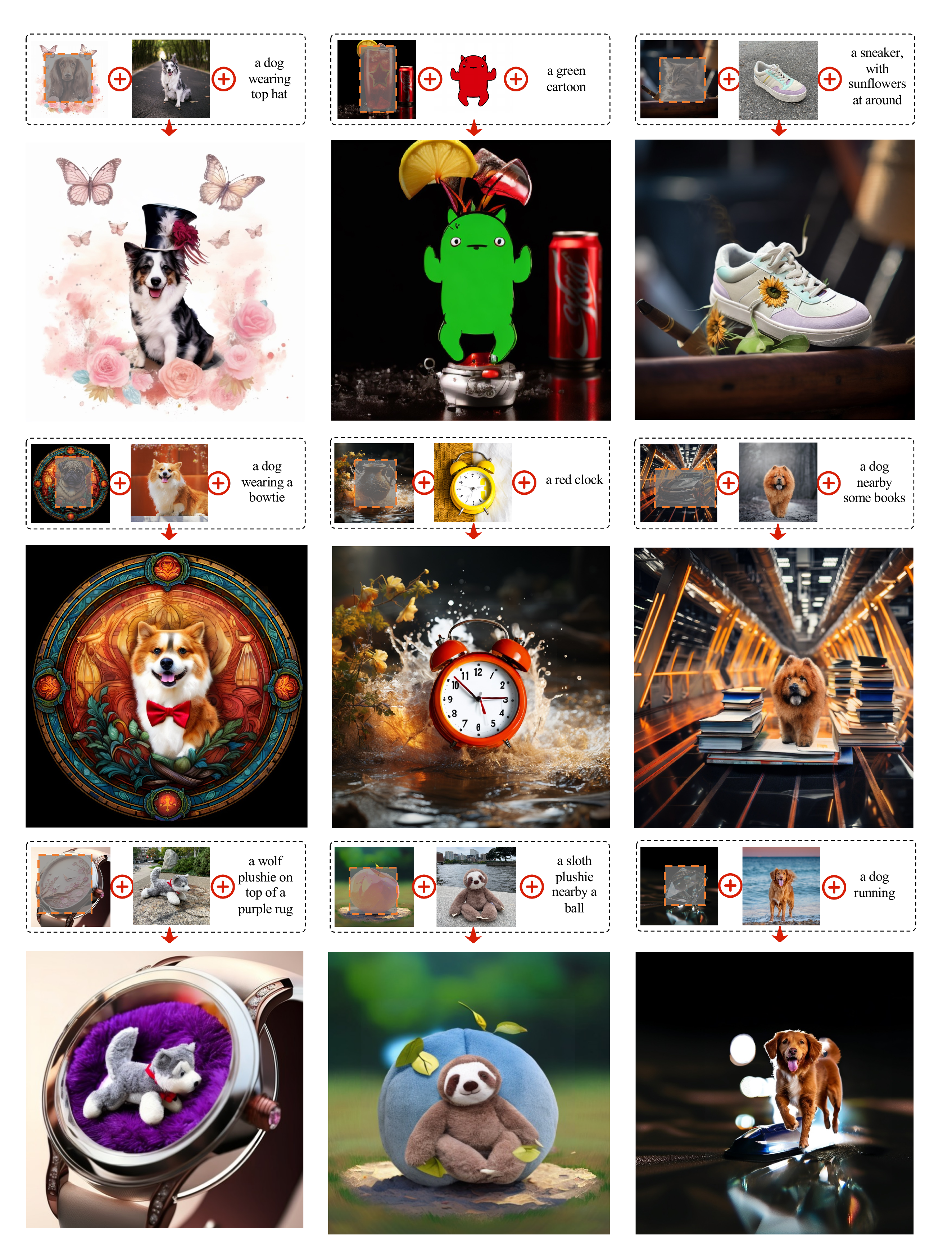}
  \caption{%
    \textbf{Qualitative results of Stable Diffusion V1.5 backbone.}
  }
  \label{fig:sd15}
  \vspace{-2mm}
\end{figure}

\begin{figure}[t]
  \centering
  \includegraphics[width=\linewidth]{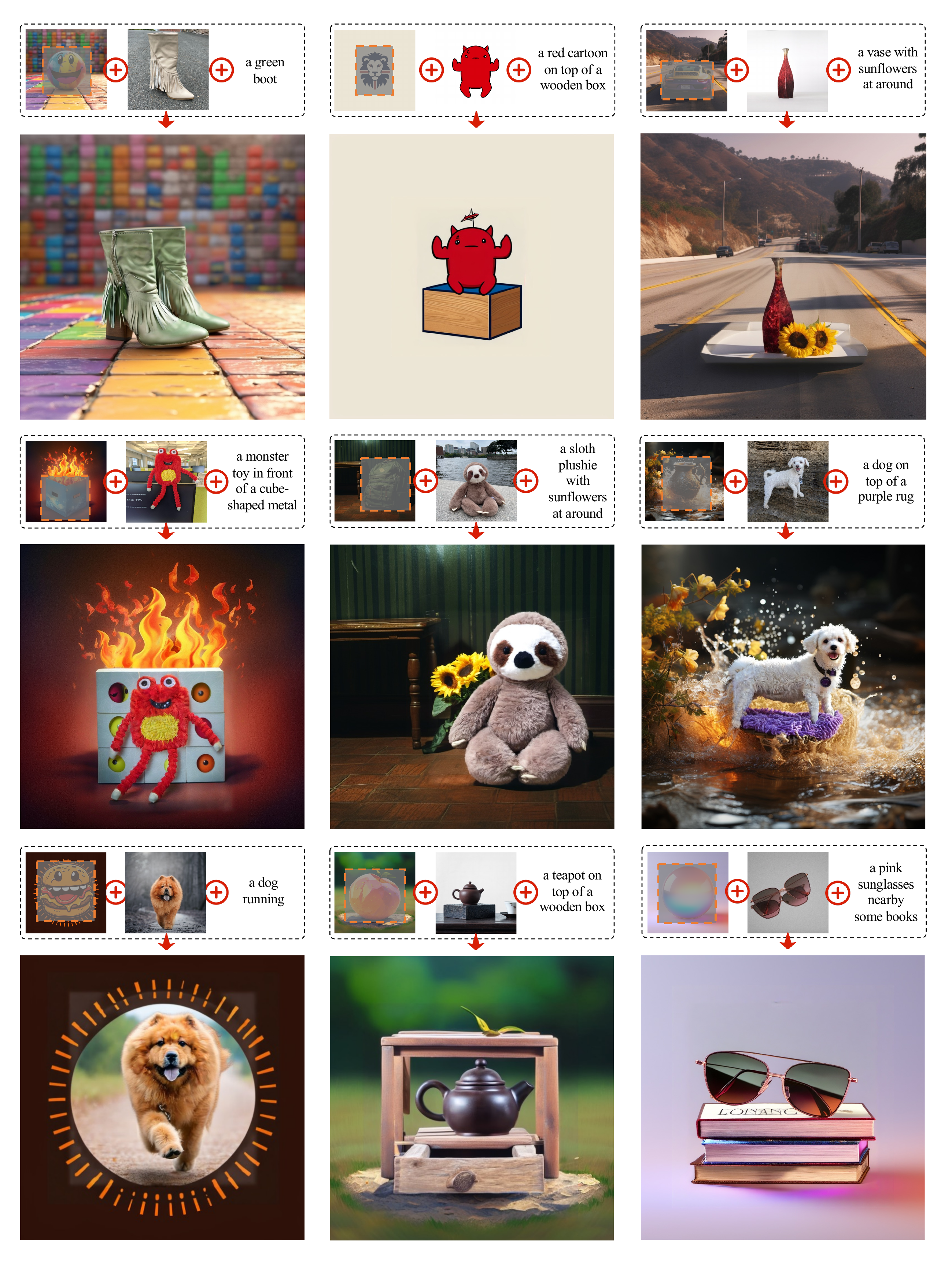}
  \caption{%
    \textbf{Qualitative results of Stable Diffusion XL backbone.}
  }
  \label{fig:sdxl}
  \vspace{-2mm}
\end{figure}

To demonstrate the generalizability of our method, we conduct experiments on Stable Diffusion V1.5\cite{sd15} and Stable Diffusion XL\cite{sdxl}, and present the qualitative results in \cref{fig:sd15} and \cref{fig:sdxl}, respectively. 
The observations indicate that our method performs well on both versions of Stable Diffusion, thereby confirming its effectiveness across different model architectures.

\clearpage
\input{sections/6.ref}